# STRUCTURING ONTOLOGIES IN A CONTEXT OF COLLABORATIVE SYSTEM MODELLING

Romy Lynn Chaib
ITAP, INRAE, Institut Agro

Montpellier, France
E-mail: romy-lynn.chaib@inrae.fr

Rallou Thomopoulos
IATE, Univ. Montpellier, INRAE, Institut Agro

F-34060 Montpellier, France
E-mail: rallou.thomopoulos@inrae.fr

Catherine Macombe
ITAP, INRAE, Institut Agro

Montpellier, France
E-mail: catherine.macombe@inrae.fr

**ABSTRACT**

Prospective studies require discussing and collaborating with the stakeholders to create scenarios of the possible evolution of the studied value-chain. However, stakeholders don't always use the same words when referring to one idea. Constructing an ontology and homogenizing vocabularies is thus crucial to identify key variables which serve in the construction of the needed scenarios. Nevertheless, it is a very complex and time-consuming task. In this paper we present the method we used to manually build ontologies adapted to the needs of two complementary system-analysis models (namely the "Godet" and the "MyChoice" models), starting from interviews of the agri-food system's stakeholders.

## 1. INTRODUCTION : A CONTEXT OF COLLABORATIVE MODELLING

Ontologies represent the semantics used by people as well as the relationships between them (Maedche & Staab, 2000; Nebot & Berlanga, 2009). They are very important when it comes to structuring knowledge and building information models (Maedche & Staab, 2000), especially since they introduce certain standards allowing the use of formalized information and vocabularies in various studies (Nebot & Berlanga, 2009).

Using the same words to refer to one concept is essential when dealing with large and complex knowledge resources, such as agri-food value-chains. Those are complex systems, made of several stakeholders interacting with each other and with their environment (Croitoru et al., 2016). Those stakeholders can be primary matter producers, breeders, transformers, distributors, consumers, but also public and private institutions, researchers, technical centers, etc… It goes without saying that all of them have different opinions as well as divergent priorities (Handayati et al., 2015), whether it is because of their position and implication in the value-chain, their political engagements, their affiliations or their life experience. They also have various different possible ways of saying the same thing: they use different ontologies to refer to same concepts. When it comes to taking a decision concerning the value-chain, it is best we have the vision and contribution of as many different points of view as possible, thus an implication of as many types of stakeholders as possible (Mitchell et al., 1997), which does not simplify the construction of a common ontology.

The case study on which this paper is based is the prospective study done on the French pork value-chain as part of project Sentinel. Indeed, the purpose of this French National Research Agency project is to improve food chemical safety along the value-chain by introducing new screening tools. In order to ensure durable applications of those tools, their impact on the value-chain must be anticipated. Nevertheless, to be able to assess the impacts of those tools, a reference of comparison must be elaborated (Pesonen et al., 2000): it consists of the likely states of the pork value chain in the future (without the new tools being implemented). Our objective is thus to model all possible evolutions of the French pork value-chain so that we can eventually evaluate the impacts certain innovations might have on it: for that, we use prospective methods (Chaib et al., 2021). This goes beyond the scope of participatory modelling: indeed, it requires not only consulting and discussing with the stakeholders (Barré, 2000; Godet & Durance, 2001; Mermet, 2004), but collaborating with them to co-design a plausible future in order to co-decide what would be best for the value-chain. We are thus in a context of collaborative modelling as described in Basco-Carrera et al. (2017).

In project Sentinel we choose to use the French prospective Godet method in which scenarios are created based on the identification of key variables (Godet, 2008). This method was however adapted due to the sanitary context in 2020 and 2021, which partly demanded the analysis of documents related to the subject (Chaib et al., 2021). In consequence, by confronting all data sources, not only did we have different ontologies between stakeholders, but those also differed from the ontologies of written documents:

indeed, authors have time to proofread and reflect on the words used, whereas stakeholders interviewed directly only have a few seconds most of the time to think about how to say their thoughts outloud.

In this paper, we talk about how we construct ontologies manually in the adapted Godet method based on interviews and documents. However, doing so is very time consuming, and gaining time would be valuable. Plus, the final list of key variables has to be reconfirmed with the stakeholders by using the Delphi method (Chaib et al., 2021). We thought it would be interesting to see how the MyChoice multicriteria argumentation tool (Thomopoulos et al., 2020) can help alleviate the disadvantages of the adapted Godet method. It could maybe help in speeding up the process of constructing the variable ontologies. This can also ensure a complete and thourough analysis of what is being said in order to increase stakeholders' awareness of certain critical situations in agri-food value-chains. For those reasons we want to explore the complementarities and redundancies of both methods.

In Section 2 we will first discuss the inputs and outputs of both methods. In Section 3, we present the steps followed in order to construct the ontology in both methods. Then, in Section 4 we examine how the adapted Godet method is relevant to our study and how the MyChoice tool can possibly help in analyzing and confirming our results. Throughout the paper, examples of what is obtained in our study on the French pork value-chain are given.

## 2. INPUTS AND OUTPUTS OF THE MODELLING PROCESS

The main goal of a collaborative knowledge representation model is to aid stakeholders so that they can make informed decisions. In the rest of the paper, we talk about the decision concerning the choice variables in order to identify the key ones so that scenarios of the evolution of the studied value-chain can be created.

For the model to be able to help stakeholders, it first has to be constructed: for that we need inputs which are then analysed in order to have outputs. Those are used in the decision making process. In this paragraph the inputs and outputs which serve us in our study are presented.

### Inputs: data from interviews and documents

Every decision making process relies on the analysis of information sourced. In our case, whether it is for the adapted Godet method or for MyChoice, information comes from different stakeholders of the value-chain as said before. It is mainly in the form of text since semi-directive interviews are conducted and then transcribed to ensure proper analysis later on. To the interviews we added documents since during the time of the study, remote work was a necessity considering the sanitary context. Each document read was considered as an interview done (Chaib et al., 2021). In total, for project Sentinel, 21 texts were analysed (12 transcribed interviews and 9 documents).

The vocabularies and the language used in the transcriptions stem from a natural discussion with the stakeholders. Each stakeholder has a different way of seeing things, analysing and interpreting them, thus the vocabulary used from one interview to another may change even though the main idea remains the same. In addition to there being varieties between the interviews, the vocabularies also vary in the documents. This can be explained by the fact that writers have time to proof-read and homogenize their words and sentences, especially those aiming to reflect a single idea, whereas stakeholders at most have a few minutes to put clear words on the idea they want to pass on. And so the question raised is: How do we treat a rather large sample of words and phrases in order to extract a limited sample of ideas?

The ultimate aim being constructing scenarios of the possible evolution of the pork value-chain, the method chosen is the French prospective collaborative Godet method as described in Godet (2008) and Godet & Durance (2001). It has the particularity of creating scenarios no stakeholder has thought of which makes the discussion and the results more interesting. In this method the problem of homogenizing ontologies is inexistent since stakeholders themselves meet and establish consensus on the main ideas to keep in mind. However, a harder option was forcibly developed because of the Covid-19 pandemic. This leads us to the following crucial topic in our paper: the outputs.

### Outputs

The outputs obtained following the interviews are double: on one hand we have outputs by using the adapted Godet method, and on another hand, we have the outputs obtained using the MyChoice tool for multicriteria argumentation since we think this method can help in constructing the ontologies needed. Both methods were initially created with different objectives in mind: the Godet method aims to identify key variables which are used for the creation of scenarios, whereas the MyChoice tool originally serves to pinpoint what may be the strengths and weaknesses of the value-chain.

*Outputs of the adapted Godet method*

The adapted Godet method consists of extracting words and phrases (which we call criteria) from the interviews and the documents. Similar criteria are then manually grouped into concepts by following an ontology matching procedure (Thomopoulos et al., 2013): basically, words or phrases which are synonyms or refer to the same idea are grouped. Concepts referring to the same global notion or theme are then grouped into variables. Each variable can take one or more value called modality (fig 1 below).

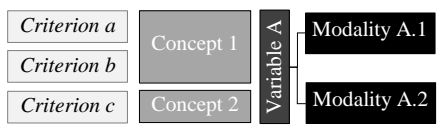

**Fig 1: Outputs obtained using the adapted Godet Method**

For example, « labour cost » and « need for investments » are concepts of the variable « production costs » which can either take the modality « production costs mastered » or the modality « fluctuating production costs ».

Depending on the explanations given during the interviews or in the documents, a concept can either be found in only one variable (which is the case for most of them) or in two variables or more. It is important to note that the variables and their modalities are identified in the list of concepts.

The identified concepts are linked to each other by influence and dependence relations identified in the transcriptions and documents and represented in mindmaps (fig 2). It is those relations which eventually help us identify key variables (Chaib et al., 2021).

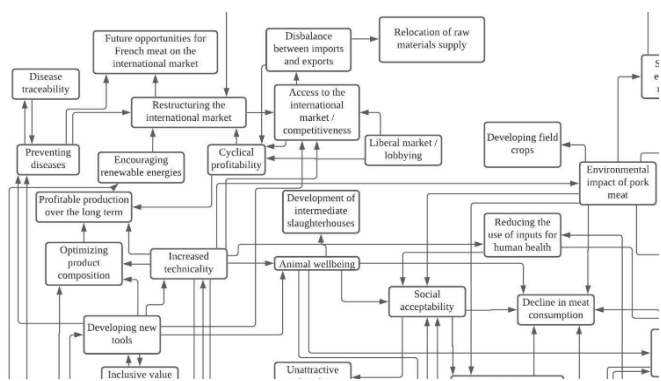

**Fig 2: Extract of a mindmap showing relations between concepts identified in an interview**

The outputs obtained then have to be confirmed by the stakeholders interviewed: a Delphi questionnaire listing all identified variables is sent to them so that they can choose 5 important variables in the 12 listed.

*Outputs of the MyChoice tool*

When using the MyChoice tool, we obtain a list of properties. Those properties are similar to what we call criteria in the adapted Godet method. Each property is attributed to an aim (which resemble the concepts of the adapted Godet method) and the aims are grouped into what is called criteria in the MyChoice tool but really is the variables of the adapted Godet method. The parallel between the denominations of each method is shown in fig 3.

There are however two main differences to note between Godet and MyChoice when it comes to the properties and the aims. The first one is that in MyChoice, a property can take several values but is still considered as a single property, whereas in the adapted Godet method we would consider that there are as many criteria as values a property can take. The second difference is that each aim can only be attributed to one single criterion, when in the adapted Godet method, a concept can be attributed to one, two or more variables.

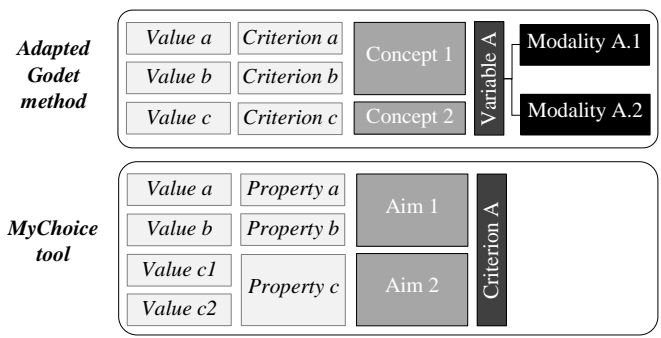

**Fig 3: Similarities in nomenclatures between the adapted Godet method and the MyChoice tool**

In addition to identifying criteria, aims and properties, the MyChoice tool helps quantify the attitude of a stakeholder concerning the alternative chosen (Thomopoulos et al., 2020). For project Sentinel, since our aim is to anticipate future evolutions of the pork value-chain, the alternative chosen is 'pursuing business as usual'.

The attitude -also called 'degree of acceptability of the alternative'- is a value between 0 and 1. It reflects to what extent pursuing business as usual meets the aims a stakeholder expressed (Thomopoulos et al., 2020). MyChoice can either give the global attitude or a stakeholder's specific attitude towards a single criterion or aim.

The following section shows how we go from the inputs to the outputs whether using the adapted Godet method or the multicriteria argumentation tool MyChoice.

## 3. FROM INPUTS TO OUTPUTS, THE ONTOLOGY-BUILDING STEPS

We saw in the previous section that the inputs for Godet and MyChoice are the same but the outputs are somewhat different. As for the processes followed: both methods are basically made of 3 main steps allowing us to build lists of variables/ criteria as shown in figure 4.

In the adapted Godet method, it is best if all of the interviews are finished so that the process of merging similar criteria is a bit easier since it is done by hand. In the MyChoice tool, homogenizing the vocabularies used is a bit easier since the aims entered in the tool are automatically registered for future choices. Nevertheless, the process of attributing aims and criteria to properties is not automated.

## 4. ALIGNMENT OF BOTH MODELS

The objective of the study is to build an ontology of variables -or criteria as they are called in the MyChoice tool- which influence the future of the value-chain and depend on it, so that we can eventually create the scenarios needed.

Using only the adapted Godet method, we were able to identify key variables and construct reference scenarios of the possible evolution of the value-chain. Nevertheless, the process of this method is very time consuming and complex as we said previously. That is why we thought about using the MyChoice tool for multicriteria argumentation. In this section we compare the results obtained using both methods to see how they can be aligned when it comes to constructing the ontologies of variables.

To simplify the analysis, in the rest of the paper we adopt the nomenclatures of the adapted Godet method. Table 1 shows the number of criteria, concepts and variables obtained using the adapted Godet method and the MyChoice tool.

The differences in number of criteria, concepts and variables can be explained as such:

| Method / tool | Criteria | Concepts | Variables | Key variables |
|---|---|---|---|---|
| Adapted Godet method | 626 | 169 | 12 | 4 |
| MyChoice | 313 | 237 | 16 | ? |

**Table 1: possible combination of MyChoice and Godet**

- For the criteria: in the adapted Godet method, those are words or phrases extracted as is from the interviews and the documents. As for the criteria (argument) in MyChoice, they are a bit more general since a phrase from an interview is dissected into the property itself, a value attributed to it, and an evaluation (+ or – in fig 5). In other words, in MyChoice, for a same denomination of a criterion, several values and evaluations can be attributed to it; they would be considered as different criteria in the adapted Godet method.
- For the concepts: in the adapted Godet method they are more general than the ones of MyChoice. A concept in Godet contains on average 4 criteria but can contain up to 24, whereas in MyChoice a concept contains 2 criteria on average and can have up to 12.
- As for the variables, they are more specific in the MyChoice database, however, some of them can be combined and it is possible to obtain 12 variables corresponding to those in the Godet method. This is more explicit in fig 5.

The fact that MyChoice is an easy tool to use makes the process of homogenizing vocabularies and constructing ontologies of variables a bit easier: indeed, it is easier to avoid redundancies between criteria, because of the separation between values, evaluations and the denomination. We find ourselves with fewer denominations and a homogenized vocabulary. It is thus easier to group them into concepts manually. In addition, the fact that each concept can only belong to one aim forces us to be specific in the nomenclatures. The variables eventually obtained using the MyChoice tool correspond to the ones obtained by following the Godet adapted method.

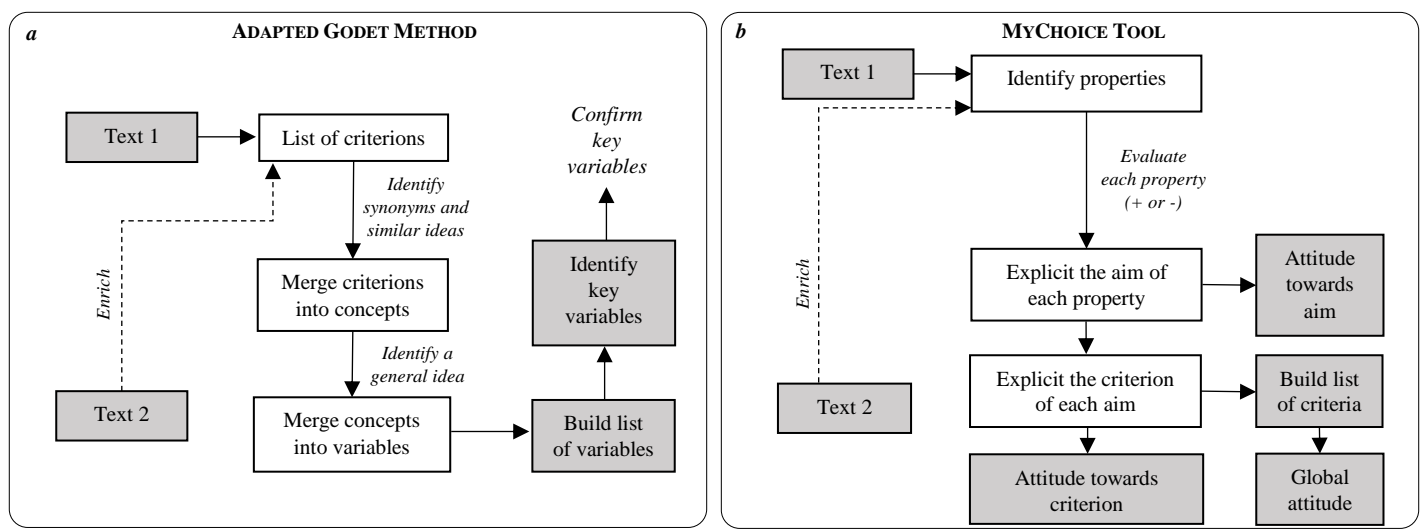

**Fig 4: Steps of building an ontology in the adapted Godet method (a) and the MyChoice tool (b) (inputs and outputs are in grey)**

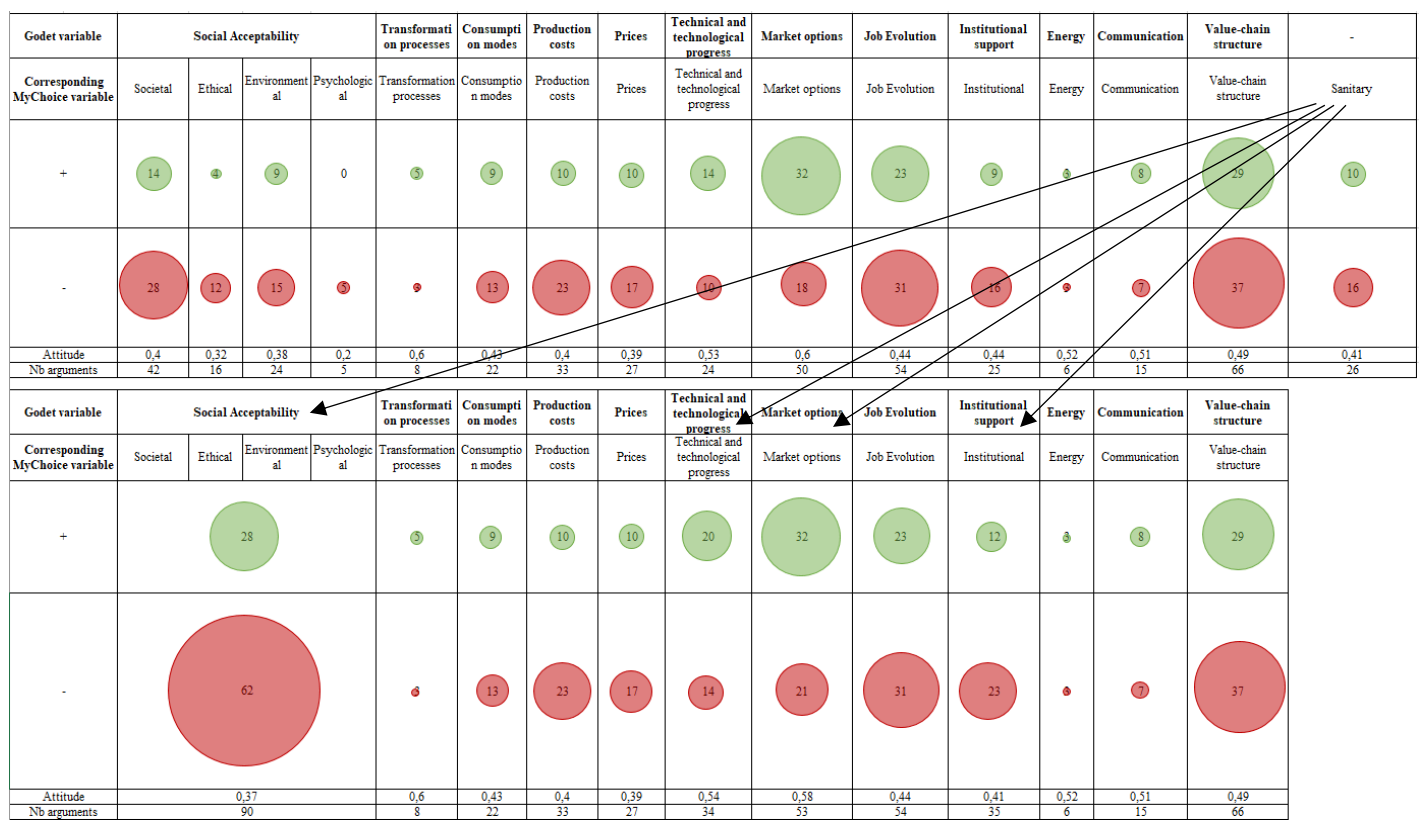

Fig 5: Identified variables in the adapted Godet method and the MyChoice tool

MyChoice thus seems to be an adapted tool for the construction of an ontology of variables: it allows us to formalize and standardize vocabularies manually but still easily compared to the adapted Godet process. This allows a better exploitation of data in the future. The MyChoice tool however does not allow us to identify key variables, at least not for now; following the process of the adapted Godet method explicited previously and detailed in Chaib et al. (2021) seems inevitable. What the MyChoice tool could facilitate is the confirmation of key variables, especially since it is sometimes rather complicated to obtain responses of stakeholders when using the Delphi method.

## 5. CONCLUSION AND FUTURE WORK

To conclude, the work done in this paper eminates from the need to identify variables and especially key variables after interviewing stakeholders and reading documents about the possible evolution of the pork value-chain taken as an example in project Sentinel. The first option explored is an adaptation of the French prospective Godet method: it consists of extracting criteria from the texts, then manually assembling them into concepts which leads to an identification of general ideas or themes we call variables. The procedure being quite heavy, we searched for alternatives that could alleviate the disadvantages of this method while also saving us time. We decided to use MyChoice since it is an easy and accessible tool: it is useful tool for the construction of ontologies since it facilitates the process, and the results attained correspond to the ones obtained using the adapted Godet method. This tool would be even more adapted and useful if the process of combining criteria and concepts was fully automated.

## ACKNOWLEDGEMENTS

This research was supported by the project SENTINEL "High-throughput screening tools for a reinforced chemical safety surveillance of food" funded by the French National Research Agency (ANR-19-CE21-0011-10).